\documentclass{article}

\usepackage{microtype}
\usepackage{graphicx}
\usepackage{subcaption}
\usepackage{booktabs} % for professional tables

\usepackage[accepted]{icml2026} 
\usepackage{hyperref}

\usepackage{icml2026}

\usepackage{amsmath}
\usepackage{amssymb}
\usepackage{mathtools}
\usepackage{amsthm}
\usepackage{colortbl}      % for \cellcolor in tables
\usepackage{enumitem}      % for itemize/enumerate spacing
\usepackage{multirow}      % for multirow in tables

\usepackage[capitalize,noabbrev]{cleveref}

\theoremstyle{plain}

\theoremstyle{definition}

\theoremstyle{remark}

\usepackage[textsize=tiny]{todonotes}
\usepackage{xcolor}
\usepackage{tcolorbox}
\usepackage{enumitem}
\icmltitlerunning{Learning Instance-Adaptive Low-Rank Orthogonal Subspaces for Clothes-Changing Person Re-Identification}

\begin{document}

\twocolumn[
  \icmltitle{Learning Instance-Adaptive Low-Rank Orthogonal Subspaces for Clothes-Changing Person Re-Identification}



  \icmlsetsymbol{equal}{*}

  \begin{icmlauthorlist}
    \icmlauthor{Dongwoo Kim}{equal,kaist}
    \icmlauthor{Tae-Kyun Kim}{equal,kaist}
  \end{icmlauthorlist}

  \icmlaffiliation{kaist}{School of Computing, KAIST, Daejeon, South Korea}
  
  \icmlcorrespondingauthor{Dongwoo Kim}{dwkim8155@kaist.ac.kr}
  \icmlcorrespondingauthor{Tae-Kyun Kim}{kimtaekyun@kaist.ac.kr}
  \icmlkeywords{low-rank representations, subspace learning, vision-language models, person re-identification, orthogonal projection, SVD, disentangled representations}

  \vskip 0.3in
]



\printAffiliationsAndNotice{}  % no special notice (required even if empty)

% --- Main Body ---
\begin{abstract}
Clothes-changing person re-identification (CC-ReID) aims to recognize individuals despite drastic appearance changes caused by clothing variation.
While existing methods rely on adversarial learning to disentangle clothing features, we propose \textbf{Ortho-ReID}, which explicitly models a \textbf{low-rank} clothing subspace from VLM text descriptions and extracts clothing-invariant representations via direct geometric constraints.
A critical component is our transformer-based \textbf{Basis Maker}, which refines a shared, low-dimensional clothing prior into an \textbf{instance-adaptive low-rank subspace} through cross-attention with image patches, enabling robust clothing feature extraction even under varying visibility conditions.
This instance-adaptive subspace is supervised via alignment with clothing text embeddings, while identity features are extracted via a learnable projection head and geometrically constrained to be strictly orthogonal to it.
Extensive experiments demonstrate state-of-the-art performance on PRCC (+5.9\% top-1), Celeb-reID-light (+3.5\%), and LaST (+5.3\%), with competitive results on LTCC.
\end{abstract}
\section{Introduction}
\label{sec:introduction}

\begin{figure}[t]
  \centering
  \includegraphics[width=\linewidth]{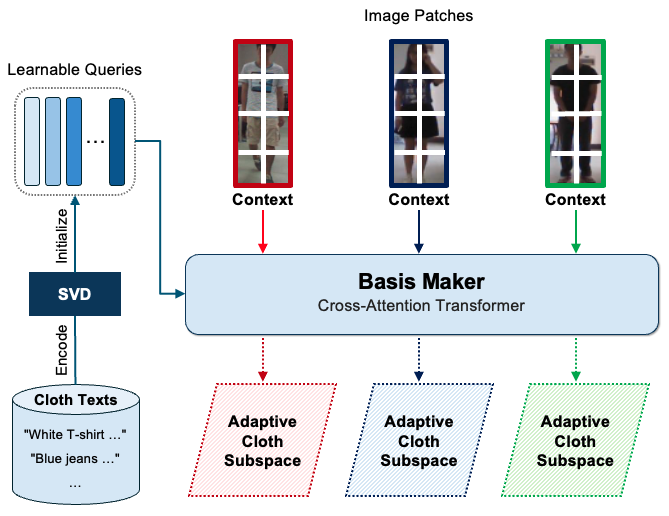}
  \caption{Instance-adaptive low-rank clothing subspace construction. Learnable queries (SVD-initialized from dataset-wide clothing text embeddings) are refined per-image via cross-attention with patch tokens into an instance-adaptive orthonormal basis spanning the low-dimensional clothing subspace.}
  \label{fig:teaser}
\end{figure}

Recent representation learning research demonstrates that high-level semantic concepts, such as ``clothing'' or ``color'', correspond to structured, low-dimensional linear subspaces within the embedding spaces of large pretrained Vision-Language Models (VLMs)~\citep{bhalla2024splice,marks2024geometry,huang2024lp++}.
This low-rank geometric structure offers a principled handle for \emph{disentangled} representation learning. 
By explicitly modeling a concept as a low-rank subspace, it becomes possible to apply geometric constraints, such as orthogonal projection, to suppress or amplify that concept in downstream representations.

We instantiate this insight in the challenging task of Clothes-Changing Person Re-Identification (CC-ReID)~\citep{wan2020person}. 
This task requires recognizing individuals across camera views despite drastic clothing variations, a scenario where appearance becomes an unreliable visual cue.
Existing methods address this through adversarial disentanglement~\citep{DIFFER_ref,gu2022clothes,xu2021adversarial}, causality-based debiasing~\citep{yang2023good}, or architectural modifications~\citep{Colors_ref}.
However, most existing approaches do not explicitly exploit the low-rank structure of VLM representations. 
Instead, they rely on learned linear projections or adversarial objectives that often lack geometric interpretability.

\begin{figure*}[!t]
\centering
\includegraphics[width=\textwidth]{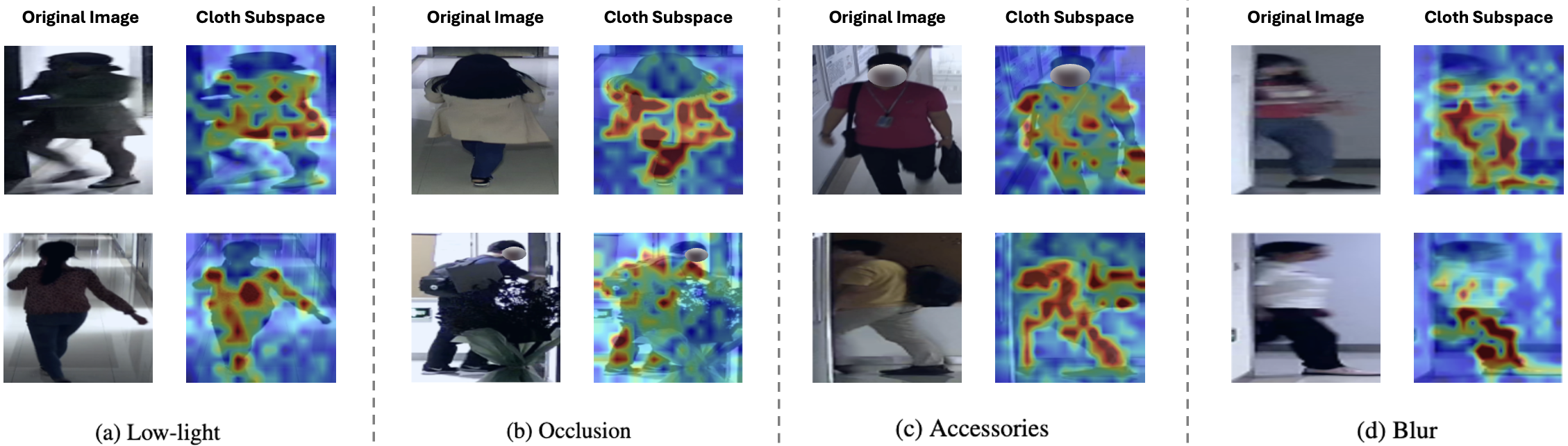}
\caption{Visualizing the Basis Maker's spatial attention under challenging conditions in the LTCC gallery. Despite severe variations in (a)~lighting, (b)~occlusions, (c)~accessories, and (d)~blur, the instance-adaptive mechanism successfully and naturally focuses on the visible clothing regions.}
\label{fig:attention}
\end{figure*}

We propose \textbf{Ortho-ReID}, a framework that treats the clothing concept as a learnable, instance-adaptive low-rank subspace and enforces identity features to be geometrically orthogonal to it.
Our approach leverages two key properties of VLM representations. 
First, clothing semantics can be captured by the principal components of CLIP text embeddings. 
We obtain these components via Singular Value Decomposition (SVD) of dataset-wide clothing descriptions, which provides a semantically grounded prior over the clothing subspace. 
Second, this global prior can be adaptively refined per-instance via cross-attention with local image patch tokens. 
This refinement yields a instance-adaptive low-rank model of the clothing in each query image.

Concretely, our transformer-based \textbf{Basis Maker} takes SVD-initialized learnable queries and attends to patch tokens to output an orthonormal basis for the clothing subspace via QR decomposition.
The identity representation is then constrained to be orthogonal to this basis during training. 
This ensures an effective geometric separation of clothing and identity information without relying on adversarial objectives.
The most closely related prior work is DIFFER~\citep{DIFFER_ref}, which also uses frozen CLIP text encoders for clothing supervision, but relies on adversarial learning rather than explicit low-rank subspace modeling.
\section{Related Work}
\label{sec:related}

\begin{figure*}[!t]
    \centering
    \includegraphics[width=\textwidth]{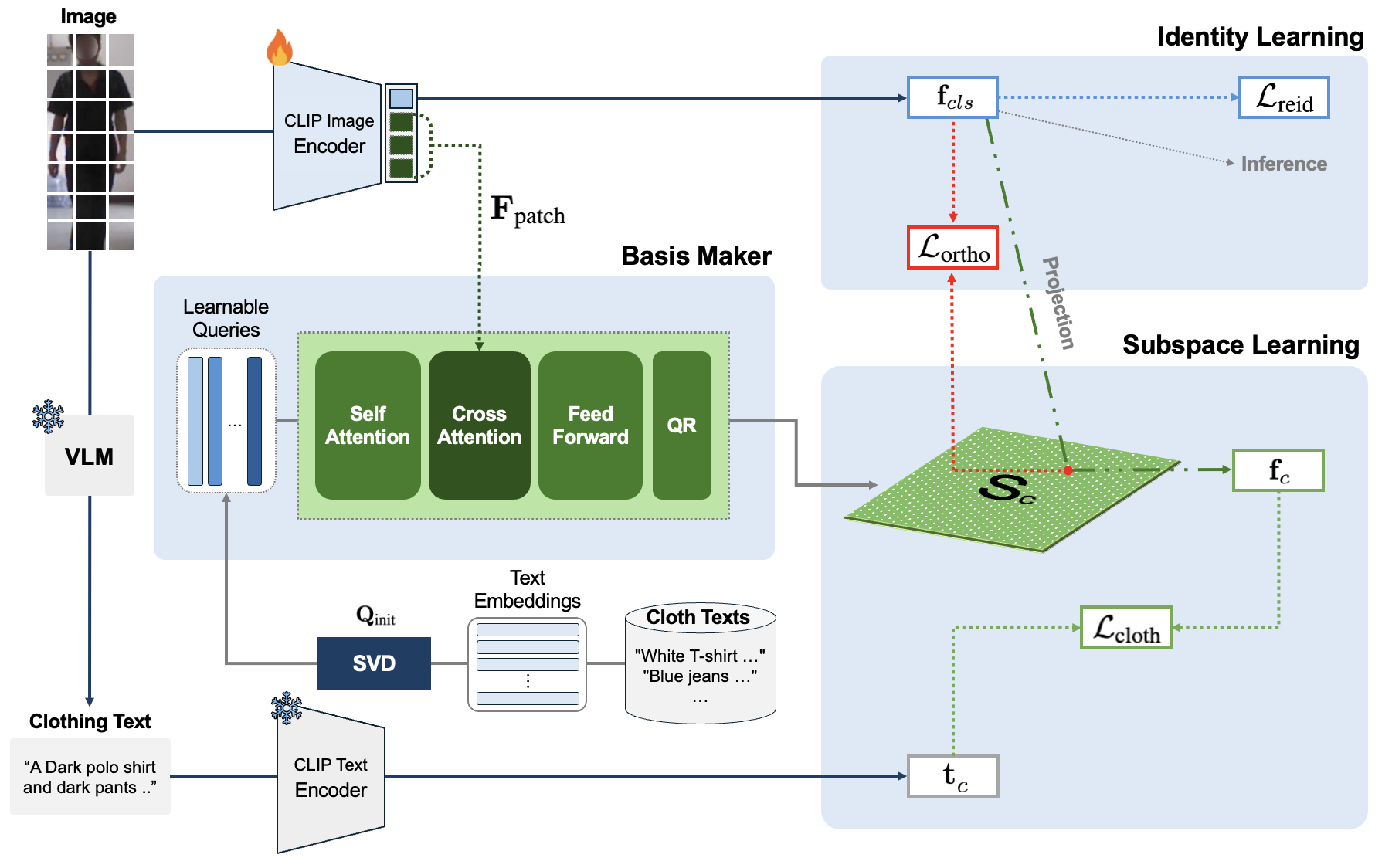}
    \caption{Overview of Ortho-ReID. The CLIP image encoder produces patch features $\mathbf{F}_{\text{patch}}$ and global feature $\mathbf{f}_{\text{cls}}$. Learnable queries are initialized via SVD of dataset-wide clothing text embeddings ($\mathbf{Q}_{\text{init}}$) and refined by the Basis Maker through cross-attention with $\mathbf{F}_{\text{patch}}$. The refined queries undergo QR decomposition to form an orthonormal basis spanning the clothing subspace $\mathbf{S}_c$. Three losses are used: $\mathcal{L}_{\text{cloth}}$ aligns projected features $\mathbf{f}_{c}$ with clothing text $\mathbf{t}_c$, $\mathcal{L}_{\text{reid}}$ for identity discrimination, and $\mathcal{L}_{\text{ortho}}$ enforces geometric orthogonality to remove clothing from $\mathbf{f}_{\text{cls}}$.}
    \label{fig:overview}
\end{figure*}

\subsection{Clothes-Changing Person Re-Identification}
CC-ReID addresses recognizing individuals despite drastic changes in apparel.
Existing approaches fall into two streams.
The first, clothes-invariant feature extraction, leverages external modalities such as 3D body shape~\citep{chen2021learning,liu2023learning}, gait~\citep{jin2022cloth}, or pose~\citep{qian2020long}.
The second, feature disentanglement, employs adversarial learning~\citep{xu2021adversarial,gu2022clothes}, causality-inspired debiasing~\citep{yang2023good}, and architectural modifications~\citep{Colors_ref}.
Recently, text-guided strategies have emerged, leveraging CLIP-based priors~\citep{li2023clip} or explicit textual supervision~\citep{DIFFER_ref,he2024instruct,li2024clip}.
Despite these advances, existing methods lack effective mechanisms for learning explicit, instance-adaptive clothing models.

\subsection{Linear Concept Representations in VLMs}
Recent studies demonstrate that high-level concepts can be approximated as linear directions or low-dimensional subspaces in large pretrained models~\citep{bhalla2024splice,huang2024lp++,marks2024geometry}.
With vision-language models, concept-aligned directions in image representations can be derived directly from textual descriptions~\citep{moayeri2023text2concept,yang2023language,bhalla2024splice}.
Recent VLM-based CC-ReID methods~\citep{DIFFER_ref,li2024clip} employ linear projections to model clothing subspaces, but construct them using simple linear matrices.
We introduce a cross-attention based Basis Maker for more flexible and expressive instance-adaptive subspace modeling.

\subsection{Orthogonality for Re-ID Disentanglement}
Projection-based methods~\citep{ravfogel2020null,ravfogel2022linear,belrose2023leace} apply fixed orthogonal projections post-hoc.
Enforcing orthogonality via in-training constraints has proven effective for disentangled feature learning.
Prior ReID work applies orthogonality to separate shape~\citep{Feng_2023_CVPR} or color~\citep{Colors_ref} from identity.
We adopt a similar training-time geometric constraint but apply it to a learned, text-grounded, instance-adaptive clothing subspace.

\section{Methodology}
\label{sec:method}

Our framework has three main components: (1) a CLIP image encoder extracting global CLS token $\mathbf{f}_{\text{cls}}$ and local patch tokens $\mathbf{F}_{\text{patch}}$, (2) a Basis Maker learning instance-adaptive clothing subspaces via cross-attention, and (3) orthogonal projection for disentanglement.
The Basis Maker uses learnable queries initialized from SVD of clothing text embeddings, refined via self-attention and cross-attention with $\mathbf{F}_{\text{patch}}$ into orthonormal bases $\mathbf{G}$ spanning $\mathcal{S}_c$.
The Basis Maker is trained by projecting $\mathbf{f}_{\text{cls}}$ onto $\mathcal{S}_c$ to extract $\mathbf{f}_{c} = \mathbf{G}\mathbf{G}^\top\mathbf{f}_{\text{cls}}$ and aligning it with clothing text embeddings via $\mathcal{L}_{\text{cloth}}$.
Clothing-invariant identity features are obtained via orthogonality loss $\mathcal{L}_{\text{ortho}}$, which trains the image encoder to produce features orthogonal to the clothing subspace.

\subsection{CLIP Encoders}
We employ EVA-02-CLIP~\citep{fang2024eva} as the image encoder, processing input image $I \in \mathbb{R}^{H \times W \times 3}$ to produce $F = [\mathbf{f}_{cls}, \mathbf{f}_1, \dots, \mathbf{f}_N] \in \mathbb{R}^{(N+1) \times d}$.
The global CLS token $\mathbf{f}_{cls}$ serves as the primary feature for the final ReID task, while local patch tokens $\mathbf{F}_{\text{patch}} = \{\mathbf{f}_i\}_{i=1}^N$ provide rich visual details for instance-adaptive subspace learning.
The text encoder remains frozen to maintain stable semantic references.

\subsection{Basis Maker}
\subsubsection{Architecture}
The Basis Maker employs a standard Transformer decoder~\citep{vaswani2017attention} with $K$ learnable queries $\mathbf{Q} \in \mathbb{R}^{K \times d}$ serving as initial basis vectors.
Inspired by Q-Former~\citep{li2023blip}, these queries dynamically extract task-relevant features through interaction with visual inputs.
To specifically adapt this architecture for modeling a strict geometric clothing subspace, we introduce SVD initialization to provide a semantic prior and QR decomposition to ensure geometric independence.
The Basis Maker processes queries through alternating self-attention and cross-attention layers.
Queries first undergo self-attention for information exchange among bases, then attend to $\mathbf{F}_{\text{patch}}$ through multi-head cross-attention.
Using patch tokens retains fine-grained spatial locality, enabling the model to capture clothing characteristics across body regions.
After feed-forward networks, the refined queries $\mathbf{Q}' \in \mathbb{R}^{K \times d}$ are orthonormalized via QR decomposition:
\begin{equation}
\mathbf{G} = \text{QR}(\mathbf{Q'}^\top) \quad \text{s.t.} \quad \mathbf{G}^\top \mathbf{G} = \mathbf{I}
\end{equation}
where $\mathbf{G} \in \mathbb{R}^{d \times K}$ forms an orthonormal basis set for the clothing subspace $\mathcal{S}_c$.
This orthonormality constraint encourages each basis vector to capture independent, non-overlapping clothing attributes (e.g., color, texture, style), preventing redundancy.

\subsubsection{Text-Guided Query Initialization}
To leverage CLIP's semantically aligned text-image space, we initialize the basis learnable queries with clothing text embeddings.
We generate natural language descriptions for each dataset using a Vision-Language Model (GPT-4o by default, open-source models yield equivalent performance; see Appendix~\ref{sec:app_vlm}) and encode them with the frozen CLIP text encoder to obtain $\{\mathbf{t}_c^{(i)}\}_{i=1}^M \in \mathbb{R}^{d}$.
We apply SVD to extract the dominant semantic directions and initialize the learnable queries with the top-$K$ principal components:
\begin{equation}
\mathbf{Q}_{\text{init}} = \text{SVD}_K(\{\mathbf{t}_c^{(i)}\}_{i=1}^M)
\end{equation}
This initialization provides a strong semantic prior ensuring queries reside in CLIP's meaningful space and attend to clothing-relevant features.
Importantly, queries remain fully learnable during training, adapting from their text-derived initialization to capture instance-specific visual clothing patterns through self-attention and cross-attention with image patch tokens.

\subsubsection{Text-Guided Semantic Alignment}
While the Basis Maker learns the clothing subspace through interaction with patch tokens, the ReID task relies on the CLS token as the global image representation.
To ensure the learned clothing subspace disentangles clothing features from the CLS token, we project $\mathbf{f}_{\text{cls}}$ onto $\mathcal{S}_c$:
\begin{equation}
\mathbf{f}_{c} = \mathbf{G} \mathbf{G}^\top \mathbf{f}_{\text{cls}}
\end{equation}
Critically, we \emph{detach} gradients of $\mathbf{f}_{\text{cls}}$ when computing this projection, ensuring only the Basis Maker parameters are updated---not the image encoder.
This gradient isolation means the Basis Maker is trained to capture clothing information as represented by the \emph{current} image encoder, while the image encoder is separately encouraged to remove clothing via $\mathcal{L}_{\text{ortho}}$.
We then align projected features with clothing text embeddings through contrastive learning:
\begin{equation}
\mathcal{L}_{\text{cloth}} = -\frac{1}{B}\sum_{i=1}^{B} \log\frac{\exp(\cos(\mathbf{f}_c^{(i)}, \mathbf{t}_c^{(i)}) / \tau)}{\sum_{j=1}^{B} \exp(\cos(\mathbf{f}_c^{(i)}, \mathbf{t}_c^{(j)}) / \tau)}
\end{equation}

\subsection{Orthogonal Projection for Disentanglement}
With the clothing subspace $\mathcal{S}_c$ learned by the Basis Maker, we apply orthogonal projection to separate clothing from identity.
We obtain clothing-invariant identity features by minimizing the similarity between $\mathbf{f}_{\text{cls}}$ and the clothing component $\mathbf{f}_{c}$ (with detached basis $\mathbf{G}$):
\begin{equation}
\mathcal{L}_{\text{ortho}} = \left\langle \frac{\mathbf{f}_{\text{cls}}}{||\mathbf{f}_{\text{cls}}||_2}, \frac{\mathbf{f}_{c}}{||\mathbf{f}_{c}||_2} \right\rangle^2
\end{equation}
By minimizing the normalized inner product, this loss encourages the image encoder to learn clothing-invariant representations during training.
At inference time, the Basis Maker is not needed and the encoder efficiently produces clothing-invariant features without additional computational overhead.

\subsection{Training Objective}
\label{sec:objective}
The total loss jointly optimizes the Basis Maker and image encoder:
\begin{equation}
\mathcal{L}_{\text{total}} = \mathcal{L}_{\text{cloth}} + \lambda_{\text{ortho}} \mathcal{L}_{\text{ortho}} + \lambda_{\text{reid}} \mathcal{L}_{\text{reid}}
\end{equation}
The ReID loss combines standard objectives with our proposed intra-modal contrastive loss:
\begin{equation}
\mathcal{L}_{\text{reid}} = \lambda_{id}\mathcal{L}_{id} + \lambda_{tri}\mathcal{L}_{tri} + \lambda_{intra}\mathcal{L}_{intra}
\end{equation}
where $\mathcal{L}_{id}$ is cross-entropy for identity prediction, $\mathcal{L}_{tri}$ is batch-hard triplet loss~\citep{hermans2017defense}, and $\mathcal{L}_{intra}$ treats all clothing variations of the same identity as uniformly positive:
\begin{equation}
\small
\mathcal{L}_{intra} = -\frac{1}{B}\sum_{i=1}^{B} \sum_{j \neq i} q_{ij} \log \frac{\exp(\mathbf{f}_{id}^{(i)} \cdot \mathbf{f}_{id}^{(j)} / \tau)}{\sum_{k \neq i} \exp(\mathbf{f}_{id}^{(i)} \cdot \mathbf{f}_{id}^{(k)} / \tau)}
\end{equation}
where $q_{ij} = \frac{1}{K_i}$ if $\text{pid}_i = \text{pid}_j$, else $0$.
This dense positive supervision is particularly valuable for CC-ReID because it simultaneously tightens intra-identity clusters despite clothing variations.
All loss weights are set to $\lambda = 1$ across all experiments.

\section{Experiments}
\label{sec:exp}

\subsection{Datasets and Metrics}
\label{sec:dataset}
We evaluate on four CC-ReID benchmarks: PRCC~\citep{yang2019prcc} (33,698 images, 221 identities, 2 outfits each), LTCC~\citep{qian2020ltcc} (17,138 images, 152 identities, avg.\ 5 clothes/person), Celeb-reID-light~\citep{huang2019celeb} (10,842 images, 590 identities), and LaST~\citep{shu2021last} (228,000 images, 10,862 identities).
We report Rank-1 accuracy and mAP.
PRCC reports Same-Clothes (SC) and Clothes-Changing (CC) results; LTCC reports CC and General protocols.

\subsection{Implementation Details}
\label{sec:impl}
We use EVA02-CLIP-L~\citep{fang2024eva} (patch size 14) as the visual encoder and GPT-4o~\citep{hurst2024gpt} to generate clothing descriptions.
Input images are resized to $224 \times 224$.
The Basis Maker comprises 6 Transformer decoder layers with 16 attention heads, and $K=16$ learnable queries initialized via SVD.
The visual encoder uses SGD ($lr = 2\!\times\!10^{-6}$, weight decay $3\!\times\!10^{-4}$); the Basis Maker uses Adam ($lr = 2\!\times\!10^{-5}$, weight decay $5\!\times\!10^{-4}$).
Both use cosine LR scheduling over 80 epochs with batch size 64 (PRCC) and 32 (others).
All experiments run on two NVIDIA A100 GPUs without data augmentation, re-ranking, or post-processing.

\subsection{Comparison with State-of-the-Art Methods}

\begin{table*}[!t]
\centering
\resizebox{\textwidth}{!}{%
\setlength\tabcolsep{5pt}
\renewcommand{\arraystretch}{1.2}
\begin{tabular}{l c c c cc cc cc cc}
\toprule
\multirow{3}{*}{\textbf{Architecture}} & \multirow{3}{*}{\textbf{Method}} & \multirow{3}{*}{\parbox{1.2cm}{\centering \textbf{Venue/}\\\textbf{Year}}} & \multirow{3}{*}{\parbox{2cm}{\centering \textbf{Additional}\\\textbf{Modality}}} & \multicolumn{4}{c}{\textbf{PRCC}} & \multicolumn{4}{c}{\textbf{LTCC}} \\
\cmidrule(lr){5-8} \cmidrule(lr){9-12}
& & & & \multicolumn{2}{c}{\textbf{CC}} & \multicolumn{2}{c}{\textbf{SC}} & \multicolumn{2}{c}{\textbf{CC}} & \multicolumn{2}{c}{\textbf{General}} \\
\cmidrule(lr){5-6} \cmidrule(lr){7-8} \cmidrule(lr){9-10} \cmidrule(lr){11-12}
& & & & \textbf{R-1} & \textbf{mAP} & \textbf{R-1} & \textbf{mAP} & \textbf{R-1} & \textbf{mAP} & \textbf{R-1} & \textbf{mAP} \\
\midrule

\multirow{5}{*}{CNN}
& CAL~\citep{gu2022clothes}        & CVPR 22 & --         & 55.2 & 55.8 & \textbf{100} & \underline{99.8} & 40.1 & 18.0 & 74.2 & 40.8 \\
& AIM~\citep{yang2023good}         & CVPR 23 & --         & 57.9 & 58.3 & \textbf{100} & \textbf{99.9} & 40.6 & 19.1 & 76.3 & 41.1 \\
& CVSL~\citep{nguyen2024cvsl}      & WACV 24 & Pose       & 57.5 & 56.9 & 97.5 & 99.1 & 44.5 & 21.3 & 76.4 & 41.9 \\
& CLIP3DReID~\citep{liu2023learning} & ICCV 23 & Text+3D  & 60.6 & 59.3 & -- & -- & 42.1 & 21.7 & -- & -- \\
& CGPG~\citep{nguyen2024cgpg}      & CVPR 24 & Silhouette & 61.8 & 58.3 & \textbf{100} & 99.6 & 46.2 & 22.9 & 77.2 & 42.9 \\
\midrule

\multirow{5}{*}{ViT}
& TransReID~\citep{he2021transreid} & ICCV 21 & --        & 46.6 & 44.8 & \textbf{100} & 99.0 & 34.4 & 17.1 & 70.4 & 37.0 \\
& CSCI~\citep{Colors_ref}           & ICCV 25 & --        & 66.2 & 61.3 & \textbf{100} & \textbf{99.9} & 47.8 & 24.4 & 82.6 & 48.0 \\
& Instruct-ReID~\citep{he2024instruct} & CVPR 24 & Text  & 54.2 & 52.3 & -- & -- & -- & -- & -- & -- \\
& MADE~\citep{peng2024masked}       & TMM 24  & Text       & 64.3 & 59.1 & \textbf{100} & 98.6 & 47.4 & 24.4 & 84.2 & 48.2 \\
& DIFFER$^\dagger$~\citep{DIFFER_ref} & CVPR 25 & Text   & \underline{68.5} & \underline{64.7} & \underline{99.9} & 99.5 & \textbf{58.2} & \textbf{31.6} & \underline{85.0} & \textbf{52.8} \\
\midrule
ViT & \textbf{Ours}                  & --      & Text       & \textbf{74.4} & \textbf{70.2} & \textbf{100} & 99.0 & \underline{56.1} & \underline{30.2} & \textbf{85.4} & \underline{52.0} \\
\bottomrule
\end{tabular}%
}
\vspace{2mm}
\caption{Comparison with state-of-the-art on PRCC and LTCC. \textbf{Bold}: best; \underline{underlined}: second-best. $^\dagger$DIFFER's LTCC result coincides with an unusually high reported baseline (54.6\%), substantially above other methods with identical settings (MADE: 45.9\%, CSCI: 44.9\%), suggesting possible implementation differences.}
\label{tab:sota_prcc_ltcc}
\end{table*}

\begin{table}[!t]
\centering
\small
\setlength\tabcolsep{6pt}
\renewcommand{\arraystretch}{1.2}
\resizebox{\columnwidth}{!}{%
\begin{tabular}{l c cc cc}
\toprule
\multirow{2}{*}{\textbf{Method}} & \multirow{2}{*}{\textbf{Venue}} & \multicolumn{2}{c}{\textbf{Celeb-L}} & \multicolumn{2}{c}{\textbf{LaST}} \\
\cmidrule(lr){3-4} \cmidrule(lr){5-6}
& & \textbf{R-1} & \textbf{mAP} & \textbf{R-1} & \textbf{mAP} \\
\midrule
CAL~\citep{gu2022clothes}    & CVPR 22 & -- & -- & 73.7 & 28.8 \\
MADE~\citep{peng2024masked}  & TMM 24  & 72.0 & 52.3 & \underline{79.0} & \underline{40.9} \\
DIFFER~\citep{DIFFER_ref}    & CVPR 25 & \underline{75.6} & \underline{54.3} & -- & -- \\
\midrule
\textbf{Ours}                & --      & \textbf{79.1} & \textbf{59.0} & \textbf{84.3} & \textbf{53.5} \\
\bottomrule
\end{tabular}%
}
\vspace{2mm}
\caption{Comparison on Celeb-reID-light (Celeb-L) and LaST.}
\label{tab:celeb_last}
\end{table}

\textbf{Results on Surveillance Datasets.}
Table~\ref{tab:sota_prcc_ltcc} shows results on PRCC and LTCC.
On PRCC, Ortho-ReID achieves 74.4\% Rank-1 and 70.2\% mAP in the CC scenario, surpassing all methods including DIFFER by +5.9\% Rank-1 and +5.5\% mAP.
On LTCC, Ortho-ReID achieves 56.1\% Rank-1 (CC), achieving highly competitive performance with DIFFER (58.2\%) while outperforming all other existing methods.

\textbf{Results on Non-Surveillance Datasets.}
Table~\ref{tab:celeb_last} shows results on Celeb-reID-light and LaST.
On Celeb-reID-light, Ortho-ReID achieves 79.1\% Rank-1 and 59.0\% mAP, outperforming DIFFER by +3.5\% and +4.7\%.
On LaST (228K identities), we obtain 84.3\% Rank-1 and 53.5\% mAP, demonstrating +5.3\% and +12.6\% improvements over MADE.
These consistent gains confirm the robustness of our approach across diverse real-world scenarios.

\subsection{Ablation Studies}

\subsubsection{Effectiveness of Orthogonal Loss}

\begin{table}[!t]
\centering
\small
\setlength\tabcolsep{4pt}
\renewcommand{\arraystretch}{1.2}
\resizebox{\columnwidth}{!}{%
\begin{tabular}{cc cc cc cc}
\toprule
\multirow{2}{*}{$\mathcal{L}_{\text{reid}}$} & \multirow{2}{*}{$\mathcal{L}_{\text{ortho}}$} & \multicolumn{2}{c}{\textbf{PRCC}} & \multicolumn{2}{c}{\textbf{LTCC}} & \multicolumn{2}{c}{\textbf{Celeb-L}} \\
\cmidrule(lr){3-4} \cmidrule(lr){5-6} \cmidrule(lr){7-8}
& & \textbf{R-1} & \textbf{mAP} & \textbf{R-1} & \textbf{mAP} & \textbf{R-1} & \textbf{mAP} \\
\midrule
\checkmark &            & 70.6 & 66.5 & 52.2 & 28.7 & 75.6 & 56.8 \\
\checkmark & \checkmark & \textbf{74.4} & \textbf{70.2} & \textbf{56.1} & \textbf{30.2} & \textbf{79.1} & \textbf{59.0} \\
\bottomrule
\end{tabular}%
}
\vspace{2mm}
\caption{Effect of orthogonal loss $\mathcal{L}_{\text{ortho}}$. Adding it consistently improves performance across all three datasets.}
\label{tab:ortho_ablation}
\end{table}

\begin{figure}[!t]
\centering
\begin{subfigure}{0.49\columnwidth}
    \centering
    \includegraphics[width=\textwidth]{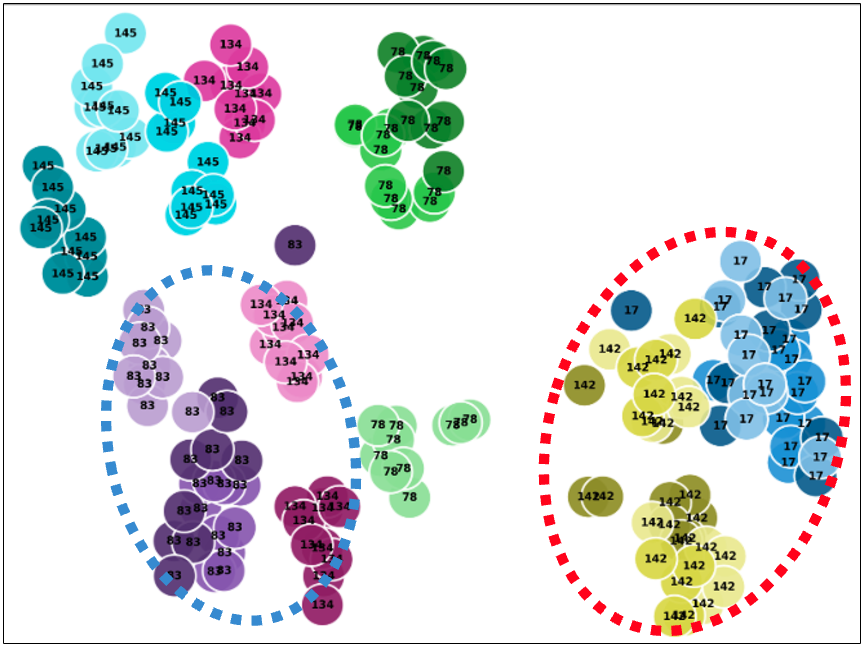}
    \caption{$\mathcal{L}_{\text{reid}}$ only}
    \label{fig:tsne_baseline}
\end{subfigure}
\hfill
\begin{subfigure}{0.49\columnwidth}
    \centering
    \includegraphics[width=\textwidth]{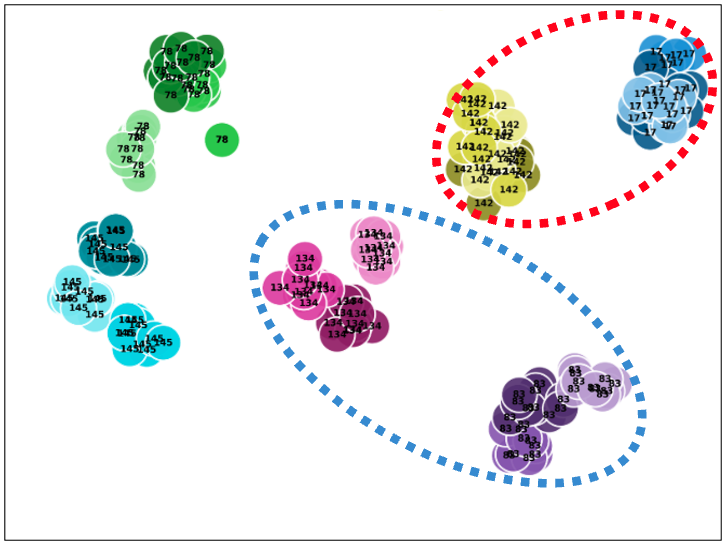}
    \caption{$\mathcal{L}_{\text{reid}}$ + $\mathcal{L}_{\text{ortho}}$}
    \label{fig:tsne_ours}
\end{subfigure}

\vspace{0.1cm}
\includegraphics[width=\columnwidth]{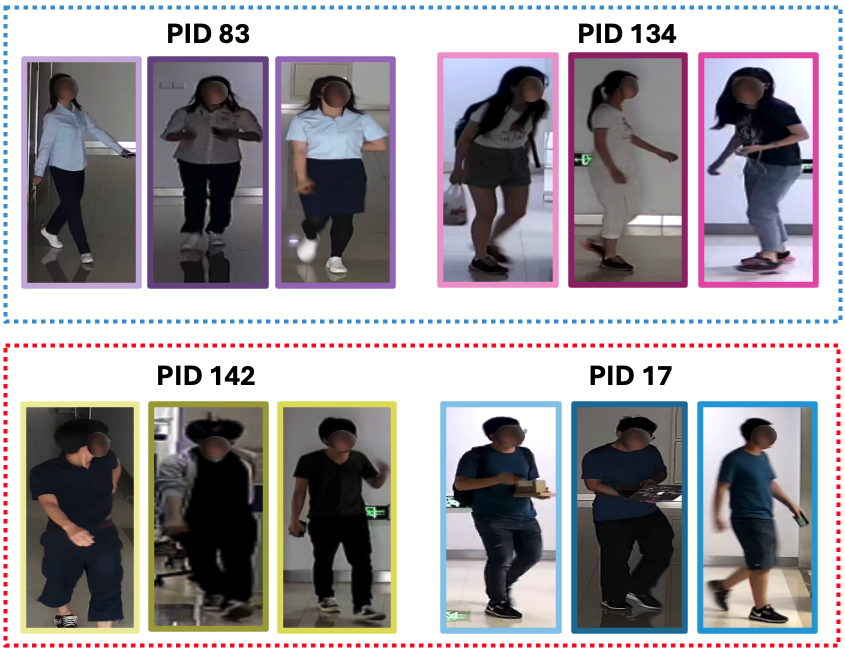}

\caption{t-SNE of learned identity features on LTCC. Colors indicate clothing; numbers denote identity. With $\mathcal{L}_{\text{ortho}}$ (right), same-identity samples cluster tightly across diverse clothing, while similar-clothing different-identity pairs are separated.}
\label{fig:tsne_comparison}
\end{figure}

Table~\ref{tab:ortho_ablation} demonstrates the critical role of the geometric constraint enforced by $\mathcal{L}_{\text{ortho}}$. 
Adding this orthogonalization step to the standard ReID baseline yields substantial and consistent Rank-1 improvements: +3.8\% on PRCC, +3.9\% on LTCC, and +3.5\% on Celeb-reID-light. This uniform gain across multiple benchmarks confirms that explicitly projecting the image embeddings away from the learned clothing subspace effectively mitigates the severe appearance bias inherent in standard baselines. 
Furthermore, Figure~\ref{fig:tsne_comparison} provides a qualitative visualization of this disentanglement effect using t-SNE. 
In the baseline model without $\mathcal{L}_{\text{ortho}}$, the feature space is heavily entangled; samples are primarily clustered by their visual appearance and clothing color, failing to group different outfits of the same person. 
In contrast, applying $\mathcal{L}_{\text{ortho}}$ fundamentally restructures the embedding space. Same-identity samples now form tight, distinct clusters regardless of their drastic clothing variations. This clearly illustrates that the geometric constraint successfully purifies the identity representation by filtering out the clothing-specific components.

\begin{table}[!t]
\centering
\footnotesize
\setlength\tabcolsep{4pt}
\renewcommand{\arraystretch}{1.2}
\begin{tabular}{l l cc cc}
\toprule
\multirow{2}{*}{\textbf{Attention Type}} & \multirow{2}{*}{\textbf{Init}} & \multicolumn{2}{c}{\textbf{PRCC}} & \multicolumn{2}{c}{\textbf{LTCC}} \\
\cmidrule(lr){3-4} \cmidrule(lr){5-6}
& & \textbf{R-1} & \textbf{mAP} & \textbf{R-1} & \textbf{mAP} \\
\midrule
\multirow{2}{*}{Self-Attn only}       & Random & 71.1 & 69.0 & 52.3 & 29.1 \\
                                       & \cellcolor{blue!10}SVD & \cellcolor{blue!10}72.9 & \cellcolor{blue!10}69.5 & \cellcolor{blue!10}54.0 & \cellcolor{blue!10}29.4 \\
\midrule
\multirow{2}{*}{Self + Cross (CLS)}   & Random & 70.4 & 67.1 & 52.5 & 28.6 \\
                                       & \cellcolor{blue!10}SVD & \cellcolor{blue!10}72.5 & \cellcolor{blue!10}69.2 & \cellcolor{blue!10}53.3 & \cellcolor{blue!10}29.8 \\
\midrule
\multirow{2}{*}{Self + Cross (Patch)} & Random & 72.3 & 69.1 & 52.5 & 29.7 \\
                                       & \cellcolor{blue!10}SVD & \cellcolor{blue!10}\textbf{74.4} & \cellcolor{blue!10}\textbf{70.2} & \cellcolor{blue!10}\textbf{56.1} & \cellcolor{blue!10}\textbf{30.2} \\
\bottomrule
\end{tabular}
\vspace{2mm}
\caption{Ablation on Basis Maker attention type and query initialization. SVD initialization (highlighted) consistently outperforms random. Patch-based cross-attention achieves best performance, with the largest gain on LTCC (+2.1\%) where clothing diversity is highest.}
\label{tab:basis_maker_ablation}
\end{table}

\subsubsection{Attention Mechanism and Initialization}
Table 4 studies the variants of attention mechanisms and query initialization in the Basis Maker. 
Patch-based cross-attention substantially outperforms both CLS-token cross-attention and self-attention alone. 
This highlights that fine-grained spatial interactions are essential for capturing locally varying clothing patterns. 
Because clothing details are distributed across different body parts and are often partially occluded, patch-level attention allows the Basis Maker to selectively focus on visible garments, yielding a +2.1\% Rank-1 gain on LTCC over the CLS variant. 
Furthermore, SVD initialization consistently surpasses random initialization across all configurations. 
This confirms the benefit of starting from a semantically meaningful prior, providing a stable optimization trajectory and ensuring the queries are properly grounded in the clothing text space from the beginning of training.

\subsubsection{Query Configuration}
Tables 5 and 6 analyze the configuration of the basis queries. 
As shown in Table 6, making the queries learnable provides a clear advantage over keeping them frozen (+2.8\% Rank-1 on LTCC). 
This demonstrates that while the SVD provides a strong global prior, allowing the queries to update during fine-tuning is crucial for adapting to the specific clothing distribution and instance-level variations of the target dataset. 
Table 5 investigates the optimal number of queries, $K$, which dictates the rank of the subspace. 
Performance peaks at $K=16$. 
A smaller $K$ restricts the representational capacity, failing to encompass the full diversity of clothing attributes. 
Conversely, a larger $K$ introduces redundancy and leads to overfitting, potentially causing the subspace to inadvertently capture identity-related or background noise.

\subsubsection{Effect of QR Decomposition}
Table 7 validates the geometric constraint imposed by QR decomposition. 
Removing this operation results in consistent performance drops across both datasets, including a notable 1.86\% decrease in LTCC mAP. 
Without QR decomposition, the dynamically generated basis vectors are prone to becoming highly correlated or collapsing into a narrower space during training. 
This confirms that explicitly enforcing orthonormality reduces redundancy among the basis vectors, ensuring a strictly structured and independent clothing subspace. 
Consequently, this geometric independence is vital for the subsequent orthogonalization step to effectively disentangle clothing features from the identity representation.

\begin{table}[!t]
\centering 
\setlength{\tabcolsep}{14pt} 
\renewcommand{\arraystretch}{1.2}
\begin{tabular}{c cc cc}
\toprule
\multirow{2}{*}{\textbf{K}} & \multicolumn{2}{c}{\textbf{PRCC}} & \multicolumn{2}{c}{\textbf{LTCC}} \\
\cmidrule(lr){2-3} \cmidrule(lr){4-5}
& \textbf{R-1} & \textbf{mAP} & \textbf{R-1} & \textbf{mAP} \\
\midrule
1  & 72.6 & 68.8 & 52.8 & 29.2 \\
4  & 72.7 & 69.4 & 54.1 & 29.8 \\
8  & 72.9 & 69.5 & 54.8 & 30.1 \\
\textbf{16} & \textbf{74.4} & \textbf{70.2} & \textbf{56.1} & \textbf{30.2} \\
32 & 73.6 & 69.4 & 55.4 & 30.3 \\
64 & 72.9 & 69.4 & 53.0 & 28.6 \\
\bottomrule
\end{tabular}
\vspace{2mm}
\caption{Ablation on the number of learnable queries $K$. Performance peaks at $K=16$; small $K$ limits coverage while large $K$ leads to overfitting.}
\label{tab:ablation_k}
\end{table}

\begin{table}[!t]
\centering
\setlength{\tabcolsep}{10pt} 
\renewcommand{\arraystretch}{1.2}
\begin{tabular}{l cc cc}
\toprule
\multirow{2}{*}{\textbf{Query Type}} & \multicolumn{2}{c}{\textbf{PRCC}} & \multicolumn{2}{c}{\textbf{LTCC}} \\
\cmidrule(lr){2-3} \cmidrule(lr){4-5}
& \textbf{R-1} & \textbf{mAP} & \textbf{R-1} & \textbf{mAP} \\
\midrule
Frozen    & 73.2 & 69.5 & 53.3 & 29.3 \\
Learnable & \textbf{74.4} & \textbf{70.2} & \textbf{56.1} & \textbf{30.2} \\
\bottomrule
\end{tabular}
\vspace{2mm}
\caption{Effect of query learnability (full Basis Maker config, SVD init). Learnable queries adapt during fine-tuning (+2.8\% on LTCC).}
\label{tab:query_learnability}
\end{table}

\begin{table}[h]
\centering
\setlength{\tabcolsep}{12pt} 
\renewcommand{\arraystretch}{1.2}
\begin{tabular}{l cc cc}
\toprule
\multirow{2}{*}{\textbf{Method}} & \multicolumn{2}{c}{\textbf{PRCC}} & \multicolumn{2}{c}{\textbf{LTCC}} \\
\cmidrule(lr){2-3} \cmidrule(lr){4-5}
& \textbf{R-1} & \textbf{mAP} & \textbf{R-1} & \textbf{mAP} \\
\midrule
w/o QR & 71.18 & 68.79 & 51.79 & 28.34 \\
w/ QR  & \textbf{74.4} & \textbf{70.2} & \textbf{56.1} & \textbf{30.2} \\
\bottomrule
\end{tabular}
\vspace{2mm}
\caption{Effect of QR decomposition. Removing it leads to consistent drops, confirming orthonormality reduces redundancy among basis vectors.}
\label{tab:validating_qr}
\end{table}

\section{Conclusion}

We present Ortho-ReID, a framework for clothes-changing person re-identification that explicitly models clothing as a text-grounded, instance-adaptive linear subspace and removes it via geometric orthogonal projection.
Our Basis Maker, initialized from SVD of clothing text embeddings and refined per-image through cross-attention with patch tokens, enables adaptive and accurate clothing feature extraction even under varied visibility conditions.
The orthogonality loss then trains the image encoder to produce representations that are geometrically decoupled from the clothing subspace, leading to consistent state-of-the-art gains across diverse CC-ReID benchmarks.

A current limitation is that our method focuses exclusively on clothing invariance: errors arise when different individuals share highly similar non-clothing attributes such as body shape, hairstyle, or posture.
Future work should extend the subspace disentanglement framework to jointly model and remove multiple distractor attributes beyond clothing.

% --- Impact Statement (required by ICML, not counted toward page limit) ---
\section*{Impact Statement}
This paper presents work whose goal is to advance the field of Machine Learning, specifically reliable long-term person re-identification under clothing variation. The primary societal application is surveillance and public safety. As with all person re-identification research, care must be taken to ensure such systems are not deployed in ways that violate privacy rights or enable mass surveillance without appropriate oversight. We do not anticipate unique ethical concerns beyond those standard to the computer vision and person re-identification research community.

% --- References (not counted toward page limit) ---
\bibliography{main}
\bibliographystyle{icml2026}

% --- Appendix (not counted toward page limit) ---
\newpage
\appendix
%% ICML 2026 Workshop Appendix for Ortho-ReID
%% Two-column layout (matching main body)

%%%%%%%%%%%%%%%%%%%%%%%%%%%%%%%%%%%%%%%%%%%%%%%%%%%%%

\section{Clothing Text Description Generation}
\label{sec:supp_clothing}
We use GPT-4o~\cite{hurst2024gpt} to generate structured clothing descriptions for each person image. The prompt below, developed with GPT-4o's assistance, instructs the model to extract clothing attributes in a consistent format. We apply this prompt to generate clothing text descriptions for all images in the PRCC~\cite{yang2019prcc}, LTCC~\cite{qian2020ltcc}, Celeb-reID-light~\cite{huang2019celeb}, and LaST~\cite{shu2021last} datasets.

\vspace{0.3cm}
\noindent\textbf{Prompt:}

\begin{tcolorbox}[colback=gray!5, colframe=gray!40, 
                  boxrule=0.5pt, arc=2pt,
                  left=2pt, right=2pt, top=2pt, bottom=2pt,
                  width=\linewidth]
{\scriptsize
\begin{verbatim}

You are given an image of an individual. 
Your task is to provide a structured 
clothing and accessories description.


Instructions:


Clothing & Accessories:
- Upper Garment: type, color, style
- Lower Garment: type, color, style  
- Footwear: type and color
- Accessories: visible items or 
  "None observed"


Clothing & Accessories sentence:
One concise sentence describing what 
the person is wearing.


Output Format:

Clothing & Accessories
- Upper Garment: [description]
- Lower Garment: [description]
- Footwear: [description]
- Accessories: [description]


Clothing & Accessories sentence: 
[Summary sentence]

\end{verbatim}
}
\end{tcolorbox}

\vspace{0.3cm}
\noindent\textbf{Example Output:}

\begin{tcolorbox}[colback=blue!5, colframe=blue!40, 
                  boxrule=0.5pt, arc=2pt,
                  left=4pt, right=4pt, top=4pt, bottom=4pt,
                  width=\linewidth]
\textbf{Clothing \& Accessories}
\begin{itemize}[leftmargin=*, topsep=2pt, itemsep=1pt]
    \item Upper Garment: Dark long-sleeve shirt over a light T-shirt
    \item Lower Garment: Dark pants
    \item Footwear: Dark shoes
    \item Accessories: None observed
\end{itemize}

\vspace{2pt}
\noindent\textbf{Clothing \& Accessories sentence:} Wearing a dark long-sleeve shirt over a light T-shirt, dark pants, and dark shoes.
\end{tcolorbox}

%%%%%%%%%%%%%%%%%%%%%%%%%%%%%%%%%%%%%%%%%%%%%%%%%%%%%

\section{VLM Comparison for Clothing Description}
\label{sec:app_vlm}

We evaluate whether the method is reproducible with open-source VLMs.
Table~\ref{tab:vlm_comparison} shows that InternVL3.5 (8B)~\citep{wang2025internvl} and Qwen3-VL (8B)~\citep{yang2025qwen} achieve performance within 0.2\% of GPT-4o, confirming that our method does not require proprietary models.

\begin{table}[h]
\centering
\renewcommand{\arraystretch}{1.2}
\resizebox{\columnwidth}{!}{% 이 부분으로 표를 감쌉니다
\begin{tabular}{lc cc cc}
\toprule
\multicolumn{2}{c}{\textbf{VLM}} & \multicolumn{2}{c}{\textbf{PRCC (CC)}} & \multicolumn{2}{c}{\textbf{LTCC (CC)}} \\
\cmidrule(lr){1-2} \cmidrule(lr){3-4} \cmidrule(lr){5-6}
\textbf{Model} & \textbf{Params} & \textbf{R-1} & \textbf{mAP} & \textbf{R-1} & \textbf{mAP} \\
\midrule
InternVL3.5~\citep{wang2025internvl} & 8B & 74.1 & 70.1 & 55.9 & \textbf{30.3} \\
Qwen3-VL~\citep{yang2025qwen}        & 8B & \textbf{74.6} & 70.0 & 56.0 & 30.0 \\
GPT-4o~\citep{hurst2024gpt}          & -- & 74.4 & \textbf{70.2} & \textbf{56.1} & 30.2 \\
\bottomrule
\end{tabular}%
}
\vspace{2mm}
\caption{Performance across VLMs for clothing description generation. Open-source 8B models match GPT-4o performance ($<$0.2\% gap).}
\label{tab:vlm_comparison}
\end{table}

%%%%%%%%%%%%%%%%%%%%%%%%%%%%%%%%%%%%%%%%%%%%%%%%%%%%%

\section{SVD-Based Query Initialization}
\label{sec:app_svd}

Algorithm~\ref{alg:svd_init} describes the SVD initialization procedure.
We use \texttt{torch.linalg.svd} for efficiency.
Embeddings are L2-normalized before concatenation, and the matrix is centered before SVD to capture variance not mean direction.
SVD is preferred over ICA or K-Means as it provides the optimal low-rank approximation aligned with our subspace learning objective.

\begin{algorithm}[h]
\caption{SVD-based Query Initialization}
\label{alg:svd_init}
\begin{algorithmic}[1]
\REQUIRE Dataset $\mathcal{D}$, CLIP text encoder $f_{\text{text}}$, query count $K$
\ENSURE $\mathbf{Q}_{\text{init}} \in \mathbb{R}^{K \times d}$
\STATE Generate clothing descriptions $\{s_i\}_{i=1}^M$ via VLM
\STATE Encode \& L2-normalize: $\mathbf{t}_c^{(i)} = f_{\text{text}}(s_i) / \|f_{\text{text}}(s_i)\|_2$
\STATE Stack: $\mathbf{T} = [\mathbf{t}_c^{(1)}, \ldots, \mathbf{t}_c^{(M)}]^\top \in \mathbb{R}^{M \times d}$
\STATE Center: $\mathbf{T}_{c} = \mathbf{T} - \text{mean}(\mathbf{T}, 0)$
\STATE SVD: $\mathbf{U}, \mathbf{\Sigma}, \mathbf{V}^\top = \text{SVD}(\mathbf{T}_{c})$
\STATE \textbf{return} $\mathbf{Q}_{\text{init}} = \mathbf{V}[:, :K]^\top$
\end{algorithmic}
\end{algorithm}

%%%%%%%%%%%%%%%%%%%%%%%%%%%%%%%%%%%%%%%%%%%%%%%%%%%%%

\end{document}